\pdfoutput=1
\documentclass[pmlr,twocolumn,10pt]{jmlr} 

\mlhtrack{findings}

\newif\iffinal
\finaltrue  

\iffinal
    \ifmlhneedspmlr
      \jmlrvolume{XXX}
    \fi
    \ifmlhfindings \jmlrproceedings{}{ML4H 2026 -- Findings Track}\fi
    \ifmlhdemo     \jmlrproceedings{}{ML4H 2026 -- Demo Track}\fi
    \jmlrworkshop{Machine Learning for Health (ML4H) 2026}
\else
    \jmlrproceedings{}{Submitted to ML4H 2026: \mlhtrackname}
    \jmlrworkshop{Machine Learning for Health (ML4H) 2026}
\fi

\makeatletter
\renewcommand*{\@titlefoot}{\scriptsize\copyright\space 2026\space\@jmlr@authors.\hfill\@reprint}
\makeatother

\usepackage{booktabs}
\usepackage{siunitx}

\title[Time-Context up to 30 Years for CRC Prediction]{How Deep is the Signal? Time-Context up to 30 Years for Colorectal Cancer Prediction in a Primary-Care EHR Model}

\author{%
 \Name{Achille Raviez\nametag{$^{1,2}$}} \Email{a.raviez@imperial.ac.uk}\\
 \Name{Lan Wang\nametag{$^{1}$}} \Email{lan.wang12@imperial.ac.uk}\\
 \Name{Brendan Delaney\nametag{$^{1}$}} \Email{brendan.delaney@imperial.ac.uk}\\
 \addr $^{1}$I-X, Imperial College London, United Kingdom\\
 \addr $^{2}$Sorbonne Universit\'e, AP-HP, France}

\begin{document}

\maketitle

\begin{abstract}
EHR foundation models read medical events within a fixed context window, and long context has so far been measured only in tokens. How far back in time a model must look to capture the earliest predictive signal is unknown. We introduce time-context, the calendar lookback window before the prediction point, and provide, to our knowledge, the first systematic evaluation of decade-scale time-context for an EHR foundation model. We pretrained a decoder-only transformer (34.6M parameters) on 1.4B coded events from 3.6M primary-care patients in North West London and fine-tuned it to predict incident colorectal cancer under a 12-month pre-diagnostic buffer. A frozen cohort of 1,008 cases and 9,582 matched controls, all registered for at least 30 years, was evaluated at 15 nested lookback windows from 2 to 30 years, so that every truncation is compared within the same patients. Discrimination rose from an AUROC of 0.551 at 2 years to 0.590 at 18 years, then plateaued, with about 80\% of the gain reached within 10 years; an XGBoost baseline converged to the same level (peak AUROC 0.589). A simulation further suggests that rotary position embeddings (RoPE) over elapsed time alias beyond about 20 years at the default base frequency, motivating a retuned $\theta$. Time-context is a measurable design axis for clinical prediction models, distinct from token-context, and should be reported.
\end{abstract}


\ifmlhneedsstatements
\paragraph*{Data and Code Availability}
The Discover dataset, a de-identified research mirror of the Whole Systems Integrated Care (WSIC) North West London database, is not publicly available; access is granted to approved researchers by the Discover-NOW Data Access Committee. Code may be available from the authors on reasonable request, subject to approval by the data controller and Imperial College London.

\paragraph*{Institutional Review Board (IRB)}
The study falls under the Discover-NOW research database ethics and HRA approval (REC 23/WM/0196) and was authorised by the North West London Data Access Committee; all analyses ran in a secure environment without export of patient-level data.
\fi

\section{Introduction}
\label{sec:intro}

\textbf{Lookback depth -- how much of a patient's past a prediction model is allowed to see -- is a weakly scrutinised design choice} \citep{pungitore2023window}, often constrained by technical limits and by shallow hospital datasets. In longitudinally sparse EHRs, calendar history is a correlated but distinct axis from the number of tokens a model reads, and it remains largely unexplored. \citet{wornow2025context} varied the context window of EHR foundation models from 512 to 16,384 tokens and showed that larger token counts improve prediction. However, a patient with dense recent contact can exhaust a large context window within a few years, while a sparsely recorded patient's whole life can fit in a few hundred tokens. The questions that matter operationally -- record retention, data curation, screening strategies, guidelines for clinical prediction models -- are all denominated in years. The median context window of EHR models has roughly quadrupled since 2019 \citep{guo2026systematic}, making long medical sequences tractable. Yet the scarcity of accessible primary-care databases, highlighted in a recent taxonomy of EHR event-stream modelling \citep{zhu2026taxonomies}, has kept researchers on secondary-care data and prevented the field from exploiting long-range signal. The same review identifies long context as an open challenge, and information sufficiency, i.e.\ how much history a reliable inference requires, as an unresolved question. Previous studies used fixed lookback windows of at most three years and did not encode event timing, using classical models \citep{hardin2021lookback} or note encoders \citep{mohammadi2026temporal}.

\textbf{The field has meanwhile moved towards generative decoder-only architectures} in addition to classical encoders, with models such as Foresight \citep{kraljevic2024foresight} and SurvivEHR \citep{gadd2026survivehr} built on causal pretraining, an approach shown to work at health-system scale \citep{rajamohan2025foundation}. Delphi-2M \citep{shmatko2025delphi} models lifetime disease trajectories but from sparser data, including registries, rather than granular clinical event streams. The closest prior work on our data source selected three-year pathways with a 1-month buffer to predict lung cancer, using an encoder-only model in the BERT lineage \citep{wang2024transformer,devlin2019bert}. Following the current state of the art, we trained a LLaMA-like decoder-only model \citep{llama2024herd}, deliberately small (34.6M parameters, against 1.6B for \citealp{rajamohan2025foundation}) to fit our timeline, compute, and training sample. 

\textbf{CRC is a well-suited probe} of how far back predictive signal carries for a specific disease in primary-care EHRs, as the adenoma--carcinoma sequence can run from about 10 to 26 years before diagnosis \citep{kuntz2011systematic}. A curve still rising at deep windows would indicate long-dwell prodromal signal of direct interest for early detection, whereas an early plateau would give a citable bound for data curation. \medskip

\textbf{Our contributions are:} (1) time-context, a design axis distinct from token-context, evaluated for the first time at decade scale through a CRC predictive-signal depth curve; (2) a within-patient nested design that isolates calendar time on a fixed cohort at fifteen windows; (3) the use of primary-care data, known to be understudied in the field, with more than 30 years of history (1990--2026); and (4) simulation results suggesting that RoPE over elapsed weeks at the default $\theta = 10{,}000$ may alias attention similarity beyond $\sim$20-year gaps, and that a retuned $\theta$ restores near-monotonic decay up to 35 years.

\section{Data and Design}
\label{sec:data}

\textbf{Data: Whole Systems Integrated Care (WSIC) North West London primary care} database. A frozen extract (24 August 2026) was restricted to events from 1 January 1990, comprising $\sim$6M patients and $\sim$2B events coded in SNOMED CT (observations) and dm+d (prescriptions).

\textbf{Incident CRC cases} were defined by a first-ever code among 73 SNOMED CT concepts descending from \emph{Malignant neoplasm of large intestine} (363510005), excluding secondary, metastatic, recurrent and non-carcinoma tumours, on or after 1 January 1990; this diagnosis date is the index date. Of 18,104 incident cases, 15,527 were registered at index, and 1,249 had at least 30 years of continuous registration before it: the deep cases (\figureref{fig:flowchart}). In each of the fine-tuning, validation and evaluation partitions, cases were matched to alive, CRC-free controls on sex, birth year ($\pm$2 years) and available continuous registration depth. Controls inherited their case's index date, and no patient appears in two partitions. Patients with fewer than five events were excluded, and realised matching ratios are reported after this threshold. The evaluation cohort is a frozen $\sim$80\% hash split of the deep cases: 1,008 cases and 9,582 matched controls (1:9.51), excluded from all preprocessing statistics and from training. The remaining 241 deep cases and all non-deep cases entered fine-tuning with full-history inputs (13,490 cases, 37,321 controls, 1:2.77), except 10\% of cases with 22--29 years of history, which formed the validation cohort (284 cases, 2,778 controls, 1:9.78).

\textbf{A uniform 12-month buffer} before the index date was removed for every patient outside pretraining. It conservatively prevents leakage of the diagnostic work-up and isolates pre-clinical, long-range signal; its length exceeds published timings of the pre-diagnostic code surge for CRC, up to 8--10 months \citep{kuiper2021steep,moullet2022prediagnostic}. The end of the buffer is the index date and prediction point.

\textbf{Evaluation patients were scored at fifteen nested time windows of 2, 4, \dots, 30 years} (\figureref{fig:design}), obtained by removing all events older than the window start, counted back from the index date. This design keeps everything constant except the depth of history: the same 10,590 patients, index dates, and the same model applied to every window. Validation patients were scored at six windows from 2 to 22 years. The primary metric is the AUROC; the AUPRC is secondary. Patients were resampled with replacement 1,000 times and each resample was scored at all fifteen windows; 95\% CIs are the 2.5th--97.5th percentiles, and between-window differences ($\Delta$AUROC) are computed within each resample, so they are paired. The plateau depth is the shallowest window whose AUROC is within 0.001 of the maximum, whose paired $\Delta$AUROC against the 30-year window has a 95\% CI containing 0, and whose point estimate of that difference lies within $\pm$0.01.

\textbf{Baseline: An XGBoost classifier} \citep{chen2016xgboost} on per-code event counts was trained using the same training data and nested evaluation as the transformer (\appendixref{apd:training}). 

\section{Model and Implementation}
\label{sec:model}

\textbf{Model: Decoder-only transformer} (causal masking, RMSNorm, SwiGLU), 34.6M parameters, $d=512$, 8 layers, 8 heads, context 8,192 tokens, full attention.

\textbf{Each event is encoded as a triplet of feature, value and time}, as reviewed by \citet{zhu2026taxonomies} and used in SurvivEHR \citep{gadd2026survivehr}. Each of the three parallel channels is projected to 512 dimensions and summed element-wise (\figureref{fig:eventstream}). The feature embedding uses a code lookup table, values use a small trained MLP over the z-score and a presence flag (plus an acute/repeat qualifier embedding for prescriptions), and time is a sinusoidal encoding of age at event (sixteen bands with periods of $\sim$1.5--630 years, \citealp{vaswani2017attention}). This tuple keeps sequences short enough for full attention and fitting within the 8,192-token context window: the median evaluation patient reaches $\sim$1,200 tokens at 30 years and the 99th percentile among deep cases $\sim$5,100.

\textbf{We apply Rotary Position Embedding (RoPE) over elapsed weeks} rather than token index \citep{su2024roformer}, as in other time-aware EHR transformers \citep{zhang2025chronoformer}. Because attention between two events depends on their time interval, irregular sampling is handled natively, and same-day events are rotation-identical. No absolute calendar date is encoded, since era-shift recognition would contaminate the depth curve; the age channel also carries temporal signal. Under the default base $\theta = 10{,}000$, RoPE aliases at decade scale: attention similarity between a reference event and one 30 years earlier rebounds above the 20-year value. Sweeping $\theta \in [10^4, 10^5]$, we select $\theta = 40{,}750$, which maximises similarity spread across 3 months--35 years subject to near-monotonic decay (\appendixref{apd:rope}).

\textbf{A [CLS] token closes every sequence}; it is embedded like other events, with its own code, no value, and age at the index date. During fine-tuning only, a dedicated learned vector is added to it. Under causal attention, its final hidden state summarises the whole history and ensures that every window contains at least one token.

\textbf{After preprocessing, our vocabulary contained 18,733 tokens}: 17,184 SNOMED CT concepts from four hierarchies (finding, procedure, observable entity, situation) plus 1,546 dm+d Virtual Therapeutic Moiety (VTM) codes from prescriptions, with [CLS]/[PAD]/[UNK] reserved. Numeric values are z-scored per code on pretraining-population statistics and winsorised at the 1st/99th percentiles; prescriptions are qualified as acute or repeat. See \appendixref{apd:preproc} for preprocessing specifications.

\textbf{Pretraining used a 70\% sample of eligible patients (3.6M patients, 1.4B events)}, with a median sequence length of 150 tokens, and excluding all downstream cohorts and patients with fewer than five events or no birth date. The trunk was first trained on half of this sample for 40,000 steps, then continued on the remaining non-overlapping half for 6,500 steps. The pretraining objective is next-event prediction (causal masking): given a patient's history to event \textit{t}, predict event \textit{t+1}. The task is fully self-supervised and disease-agnostic. See \appendixref{apd:training} for training specifications.

\textbf{We fully fine-tuned the model for CRC discrimination}, training a linear head on the final-norm [CLS] representation jointly with all trunk parameters after one frozen-trunk warm-up epoch. Hyperparameters and checkpoints were selected by mean AUROC across six validation windows. The model was trained once on full-history inputs and applied unchanged to all fifteen evaluation windows, so that they differ only in their input.

\section{Results}
\label{sec:results}

\textbf{Transformer discrimination rose with time-context up to 18 years and plateaued thereafter}, with 80\% of the total gain (0.032 of 0.040 AUROC) realised within the first 10 years (\figureref{fig:depth}, \appendixref{apd:table}).

\begin{figure}[htbp]
\floatconts
  {fig:depth}
  {\caption{Evaluation AUROC (95\% CI) by lookback window for the transformer and XGBoost.}}
  {\includegraphics[width=\linewidth]{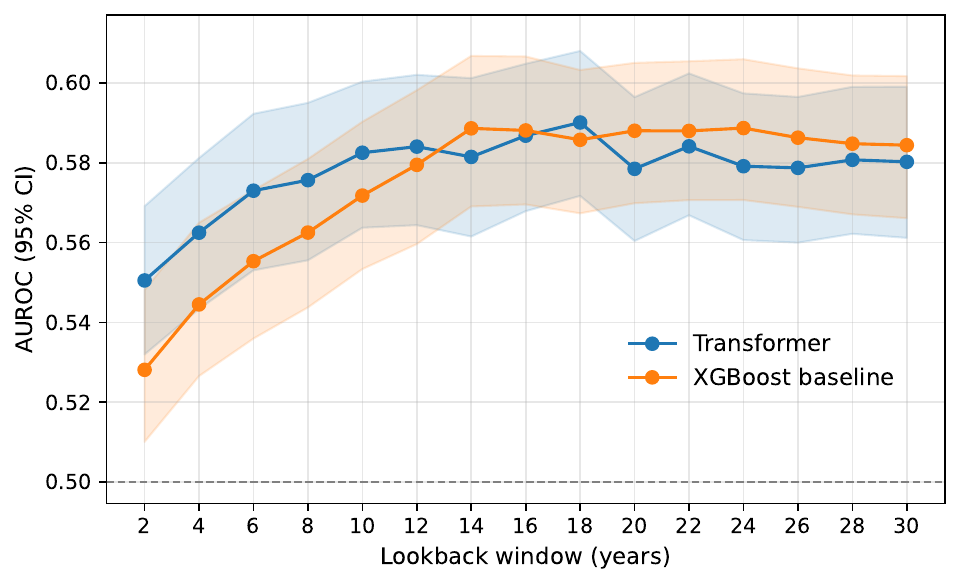}}
\end{figure}

AUROC was 0.551 (0.532--0.569) at 2 years, 0.583 (0.564--0.600) at 10 years, and reached a peak of 0.590 (0.572--0.608) at 18 years; all deeper windows stayed within $\pm$0.01 of the 30-year value (0.580, 0.561--0.599). Under the above rule, the plateau depth was 18 years: $\Delta$(30y$-$18y) $= -0.010$ ($-0.022$--$0.002$), consistent with no further gain. The shallowest window remains clearly below the plateau with $\Delta$(18y$-$2y) $= +0.040$ ($0.018$--$0.061$). AUPRC tracked the same shape, from 0.113 at 2 years to 0.132 at 18 years (prevalence 0.095).

\textbf{XGBoost on per-code counts traced a curve of similar shape with comparable performance:} AUROC 0.528 (0.510--0.548) at 2 years, rising to a peak of 0.589 (0.569--0.607) at 14 years, and flat thereafter (30 years: 0.584, 0.566--0.602). The transformer was marginally ahead at shallow windows (2-year paired $\Delta = +0.022$, $0.000$--$0.043$), but the two model classes converged from about 12 years and were statistically indistinguishable at depth (30-year paired $\Delta = -0.004$, $-0.026$--$0.017$). The transformer's advantage is concentrated where records are sparse and disappears once enough events accumulate for code counts alone to carry the signal.

\textbf{History beyond 18 years did not improve discrimination}, consistent with a saturating signal rather than an effect of diminishing record density alone. Median tokens per evaluation patient grew from $\sim$90 at 2 years to $\sim$1,200 at 30 years, although the tokens added per 2-year step fell from $\sim$180 to $\sim$10 (\figureref{fig:tokens}); between 18 and 24 years, each step still added several tens of tokens per patient without any gain. The plateau appears gradual rather than a sharp cut-off, and under data shift the curve is best read as a lower bound on the long-range signal available.

\begin{figure}[htbp]
\floatconts
  {fig:tokens}
  {\caption{Cumulative and marginal token count per lookback window for evaluation patients.}}
  {\includegraphics[width=\linewidth]{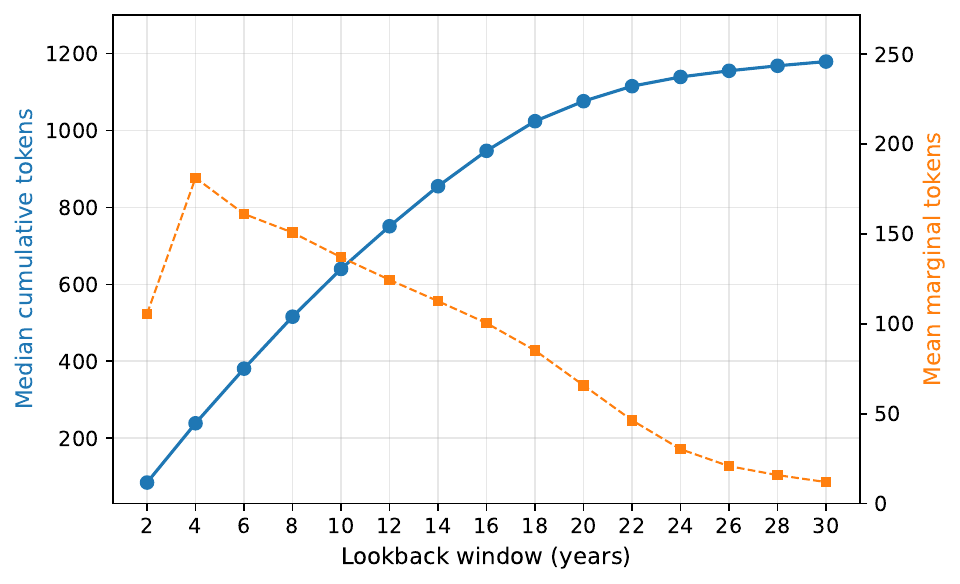}}
\end{figure}

\textbf{Absolute discrimination was modest for both models}, consistent with a conservative 12-month buffer, disease-agnostic preprocessing without cancer-specific feature selection, and noisy primary-care data. Above all, matching cases and controls on age and sex (along with registration depth) neutralises two of the strongest CRC predictors, so these figures are not comparable with clinical CRC models developed with cohort designs, such as QCancer \citep{hippisleycox2012crc}. Our 34.6M-parameter model is probably under-sized: a larger trunk, task-specific features and longer training may lift absolute discrimination; the depth shape, replicated by an independent model class, is however unlikely to change.

\section{Discussion and Conclusion}
\label{sec:discussion}

\textbf{Calendar-time lookback depth is a distinct axis from token-context length}, and it can be measured directly. For incident colorectal cancer in primary care, most of the attainable discrimination is reached by $\sim$18 years of lookback history, with no detectable gain beyond, a timescale compatible with the known adenoma-carcinoma dwell time. While specific to two model classes, a selected cohort and coded primary-care data under a 12-month buffer, these results suggest that a transformer can capture long-range pre-diagnostic signal that may reflect underlying disease biology, with early detection potential. The transformer and the count-based baseline converge in both performance and curve shape, although the transformer peaks later (18 vs 14 years), suggesting that the measured signal is largely a property of the data rather than of a specific architecture. Absolute discrimination is modest for both models, which may indicate an approximate signal ceiling for disease-agnostic primary-care codes at a buffered pre-diagnostic horizon in a matched design.

\textbf{The main limitations are:} (1) this is a single-region study, albeit across diverse London practices and populations; (2) registration depth is counted from 1990 so every deep-cohort index date falls in 2020--2026, the evaluation cohort is therefore strongly selected (older, residentially stable, likely less deprived); (3) events recorded per year increased from $\sim$5M in the 1990s to $\sim$130M in the 2020s, and the number of distinct SNOMED CT codes per year rose $\sim$2.5-fold, making older history sparser and more coarsely coded; (4) some codes have approximate dates, and registration depth was computed at year resolution, potentially blurring the 30-year window; and (5) results come from a single training run, and architectural choices need ablations.

\textbf{Future work includes} expanding downstream tasks to other tumour site such as lung and pancreas to compare depth signatures, evaluating at natural prevalence, scaling model and data, refining preprocessing, adding an explainability analysis, and comparing architectures for long-range dependencies. 
    
\acks{This work was supported by the NIHR Imperial Biomedical Research Centre (BRC) and by Discover-NOW, the health data research hub providing access to WSIC. A.R.'s visiting research position was funded by Assistance Publique -- H\^opitaux de Paris (AP-HP). The views expressed are those of the authors and not necessarily those of the NIHR or the Department of Health and Social Care.}

\bibliography{ref}

\begin{thebibliography}{21}
\providecommand{\natexlab}[1]{#1}
\providecommand{\url}[1]{\texttt{#1}}
\expandafter\ifx\csname urlstyle\endcsname\relax
  \providecommand{\doi}[1]{doi: #1}\else
  \providecommand{\doi}{doi: \begingroup \urlstyle{rm}\Url}\fi

\bibitem[Chen and Guestrin(2016)]{chen2016xgboost}
Tianqi Chen and Carlos Guestrin.
\newblock {XGBoost}: A scalable tree boosting system.
\newblock In \emph{Proceedings of the 22nd ACM SIGKDD International Conference on Knowledge Discovery and Data Mining}, pages 785--794, 2016.
\newblock \doi{10.1145/2939672.2939785}.

\bibitem[Devlin et~al.(2019)Devlin, Chang, Lee, and Toutanova]{devlin2019bert}
Jacob Devlin, Ming-Wei Chang, Kenton Lee, and Kristina Toutanova.
\newblock {BERT}: Pre-training of deep bidirectional transformers for language understanding.
\newblock In \emph{Proceedings of the 2019 Conference of the North {A}merican Chapter of the Association for Computational Linguistics: Human Language Technologies (NAACL-HLT)}, pages 4171--4186, 2019.
\newblock \doi{10.18653/v1/N19-1423}.

\bibitem[Gadd et~al.(2026)Gadd, Gokhale, Acharya, Cooper, Crowe, Fitzsimmons, Jackson, Nirantharakumar, and Yau]{gadd2026survivehr}
Charles Gadd, Krishna Gokhale, Aditya Acharya, Jennifer Cooper, Francesca Crowe, Leah Fitzsimmons, Thomas Jackson, Krishnarajah Nirantharakumar, and Christopher Yau.
\newblock {SurvivEHR}: a competing risks, time-to-event foundation model for multiple long-term conditions from primary care electronic health records.
\newblock \emph{npj Digital Medicine}, 9:\penalty0 546, 2026.
\newblock \doi{10.1038/s41746-026-02709-z}.

\bibitem[Guo et~al.(2026)Guo, Arciniegas, Yan, Fries, Tomlinson, and Sung]{guo2026systematic}
Lin~Lawrence Guo, Sofia~E. Arciniegas, Adam~P. Yan, Jason Fries, George~A. Tomlinson, and Lillian Sung.
\newblock Systematic review of foundation models for structured electronic health records.
\newblock \emph{Journal of the American Medical Informatics Association}, 33\penalty0 (6):\penalty0 1190--1198, 2026.
\newblock \doi{10.1093/jamia/ocag033}.

\bibitem[Hardin and Reps(2021)]{hardin2021lookback}
Jill Hardin and Jenna~M. Reps.
\newblock Evaluating the impact of covariate lookback times on performance of patient-level prediction models.
\newblock \emph{BMC Medical Research Methodology}, 21\penalty0 (1):\penalty0 180, 2021.
\newblock \doi{10.1186/s12874-021-01370-2}.

\bibitem[Hippisley-Cox and Coupland(2012)]{hippisleycox2012crc}
Julia Hippisley-Cox and Carol Coupland.
\newblock Identifying patients with suspected colorectal cancer in primary care: derivation and validation of an algorithm.
\newblock \emph{British Journal of General Practice}, 62\penalty0 (594):\penalty0 e29--e37, 2012.
\newblock \doi{10.3399/bjgp12X616346}.

\bibitem[Kraljevic et~al.(2024)Kraljevic, Bean, Shek, Bendayan, Hemingway, Au~Yeung, Deng, Balston, Ross, Idowu, Teo, and Dobson]{kraljevic2024foresight}
Zeljko Kraljevic, Dan Bean, Anthony Shek, Rebecca Bendayan, Harry Hemingway, Joshua Au~Yeung, Alexander Deng, Alfred Balston, Jack Ross, Esther Idowu, James~T. Teo, and Richard J.~B. Dobson.
\newblock Foresight---a generative pretrained transformer for modelling of patient timelines using electronic health records: a retrospective modelling study.
\newblock \emph{The Lancet Digital Health}, 6\penalty0 (4):\penalty0 e281--e290, 2024.
\newblock \doi{10.1016/S2589-7500(24)00025-6}.

\bibitem[Kuiper et~al.(2021)Kuiper, van Herk-Sukel, Lemmens, Kuipers, and Herings]{kuiper2021steep}
Josephina~G. Kuiper, Myrthe P.~P. van Herk-Sukel, Valery E. P.~P. Lemmens, Ernst~J. Kuipers, and Ron M.~C. Herings.
\newblock A steep increase in healthcare seeking behaviour in the last months before colorectal cancer diagnosis.
\newblock \emph{BMC Family Practice}, 22\penalty0 (1):\penalty0 121, 2021.
\newblock \doi{10.1186/s12875-021-01482-0}.

\bibitem[Kuntz et~al.(2011)Kuntz, Lansdorp-Vogelaar, Rutter, Knudsen, van Ballegooijen, Savarino, Feuer, and Zauber]{kuntz2011systematic}
Karen~M. Kuntz, Iris Lansdorp-Vogelaar, Carolyn~M. Rutter, Amy~B. Knudsen, Marjolein van Ballegooijen, James~E. Savarino, Eric~J. Feuer, and Ann~G. Zauber.
\newblock A systematic comparison of microsimulation models of colorectal cancer: the role of assumptions about adenoma progression.
\newblock \emph{Medical Decision Making}, 31\penalty0 (4):\penalty0 530--539, 2011.
\newblock \doi{10.1177/0272989X11408730}.

\bibitem[{Llama Team}(2024)]{llama2024herd}
{Llama Team}.
\newblock The {Llama} 3 herd of models.
\newblock arXiv:2407.21783, 2024.

\bibitem[Mohammadi et~al.(2026)Mohammadi, Vahdat, Jain, Namin, Palacholla, and Kamarthi]{mohammadi2026temporal}
Ramin Mohammadi, Vahab Vahdat, Sarthak Jain, Amir~T. Namin, Ramya Palacholla, and Sagar Kamarthi.
\newblock Temporal data requirement for predicting unplanned hospital readmissions.
\newblock \emph{arXiv preprint arXiv:2605.00738}, 2026.

\bibitem[Moullet et~al.(2022)Moullet, Funston, Mounce, Abel, de~Wit, Walter, and Zhou]{moullet2022prediagnostic}
Margaux Moullet, Garth Funston, Luke T.~A. Mounce, Gary~A. Abel, Niek de~Wit, Fiona~M. Walter, and Yin Zhou.
\newblock Pre-diagnostic clinical features and blood tests in patients with colorectal cancer: a retrospective linked-data study.
\newblock \emph{British Journal of General Practice}, 72\penalty0 (721):\penalty0 e556--e563, 2022.
\newblock \doi{10.3399/BJGP.2021.0563}.

\bibitem[Pungitore and Subbian(2023)]{pungitore2023window}
Sarah Pungitore and Vignesh Subbian.
\newblock Assessment of prediction tasks and time window selection in temporal modeling of electronic health record data: a systematic review.
\newblock \emph{Journal of Healthcare Informatics Research}, 7\penalty0 (3):\penalty0 313--331, 2023.
\newblock \doi{10.1007/s41666-023-00143-4}.

\bibitem[Rajamohan et~al.(2025)Rajamohan, Gao, Zhu, Huang, Chen, Cho, et~al.]{rajamohan2025foundation}
Haresh~Rengaraj Rajamohan, Xiang Gao, Weicheng Zhu, Shih-Lun Huang, Lei Chen, Kyunghyun Cho, et~al.
\newblock Foundation models for clinical records at health system scale.
\newblock \emph{arXiv preprint arXiv:2507.00574}, 2025.

\bibitem[Shmatko et~al.(2025)Shmatko, Jung, Gaurav, Brunak, Mortensen, Birney, Fitzgerald, and Gerstung]{shmatko2025delphi}
Artem Shmatko, Alexander~Wolfgang Jung, Kumar Gaurav, S{\o}ren Brunak, Laust~Hvas Mortensen, Ewan Birney, Tom Fitzgerald, and Moritz Gerstung.
\newblock Learning the natural history of human disease with generative transformers.
\newblock \emph{Nature}, 647\penalty0 (8088):\penalty0 248--256, 2025.
\newblock \doi{10.1038/s41586-025-09529-3}.

\bibitem[Su et~al.(2024)Su, Ahmed, Lu, Pan, Bo, and Liu]{su2024roformer}
Jianlin Su, Murtadha Ahmed, Yu~Lu, Shengfeng Pan, Wen Bo, and Yunfeng Liu.
\newblock {RoFormer}: Enhanced transformer with rotary position embedding.
\newblock \emph{Neurocomputing}, 568:\penalty0 127063, 2024.
\newblock \doi{10.1016/j.neucom.2023.127063}.

\bibitem[Vaswani et~al.(2017)Vaswani, Shazeer, Parmar, Uszkoreit, Jones, Gomez, Kaiser, and Polosukhin]{vaswani2017attention}
Ashish Vaswani, Noam Shazeer, Niki Parmar, Jakob Uszkoreit, Llion Jones, Aidan~N. Gomez, {\L}ukasz Kaiser, and Illia Polosukhin.
\newblock Attention is all you need.
\newblock In \emph{Advances in Neural Information Processing Systems}, volume~30, 2017.

\bibitem[Wang et~al.(2024)Wang, Yin, Glampson, Peach, Barahona, Delaney, and Mayer]{wang2024transformer}
Lan Wang, Yonghua Yin, Ben Glampson, Robert Peach, Mauricio Barahona, Brendan~C. Delaney, and Erik~K. Mayer.
\newblock Transformer-based deep learning model for the diagnosis of suspected lung cancer in primary care based on electronic health record data.
\newblock \emph{eBioMedicine}, 110:\penalty0 105442, 2024.
\newblock \doi{10.1016/j.ebiom.2024.105442}.

\bibitem[Wornow et~al.(2025)Wornow, Bedi, Fuentes~Hernandez, Steinberg, Fries, R{\'e}, Koyejo, and Shah]{wornow2025context}
Michael Wornow, Suhana Bedi, Miguel~Angel Fuentes~Hernandez, Ethan Steinberg, Jason~Alan Fries, Christopher R{\'e}, Sanmi Koyejo, and Nigam~H. Shah.
\newblock Context clues: Evaluating long context models for clinical prediction tasks on {EHR} data.
\newblock In \emph{International Conference on Learning Representations (ICLR)}, 2025.
\newblock arXiv:2412.16178.

\bibitem[Zhang and Li(2025)]{zhang2025chronoformer}
Yuanyun Zhang and Shi Li.
\newblock {ChronoFormer}: Time-aware transformer architectures for structured clinical event modeling.
\newblock \emph{arXiv preprint arXiv:2504.07373}, 2025.

\bibitem[Zhu et~al.(2026)Zhu, Liu, Luo, and Zhu]{zhu2026taxonomies}
Mingcheng Zhu, Yu~Liu, Zhiyao Luo, and Tingting Zhu.
\newblock The taxonomies, training, and applications of event stream modelling for electronic health records.
\newblock \emph{arXiv preprint arXiv:2603.14003}, 2026.

\end{thebibliography}

\appendix

\section{RoPE base-frequency retuning for long-range gaps}\label{apd:rope}

Attention similarity between two events should decay monotonically as the temporal gap increases. To verify this, we drew 20,000 random vectors in simulation (head dimension 64), applied RoPE rotations using elapsed time in weeks for gaps ranging from 3 months to 35 years, and measured the mean rotated similarity as a function of elapsed time (\figureref{fig:rope}). We represent time in weeks so that a 35-year gap spans $\sim$1,800 position units, comparable in magnitude to the token-index ranges RoPE was designed for; in days the same gap would be $\sim$12,800. Any rise in the curve indicates aliasing, i.e.\ distant events regaining spurious similarity once the faster rotations complete full turns. After sweeping $\theta \in [10^4, 10^5]$, we screened $\theta \in \{10{,}000;\ 40{,}750;\ 75{,}000;\ 100{,}000\}$ and scored each candidate by the number of non-monotonic rebounds and by the dynamic range of the similarity curve. At the default $\theta$ = 10,000, similarity rebounds beyond $\sim$20 years: events 25--35 years apart appear more alike than events 20 years apart; whereas $\theta$ = 40,750 yields near-monotonic decay with the widest dynamic range and was retained for all experiments. While this method allows us to justify our parameter choice given our long context window, further ablations are needed to demonstrate its impact on performance.

\begin{figure}[htbp]
\floatconts
  {fig:rope}
  {\caption{Simulated mean RoPE attention similarity between random query/key vectors as a function of elapsed time (0.25--35 years), for four base frequencies $\theta$.}}
  {\includegraphics[width=\linewidth]{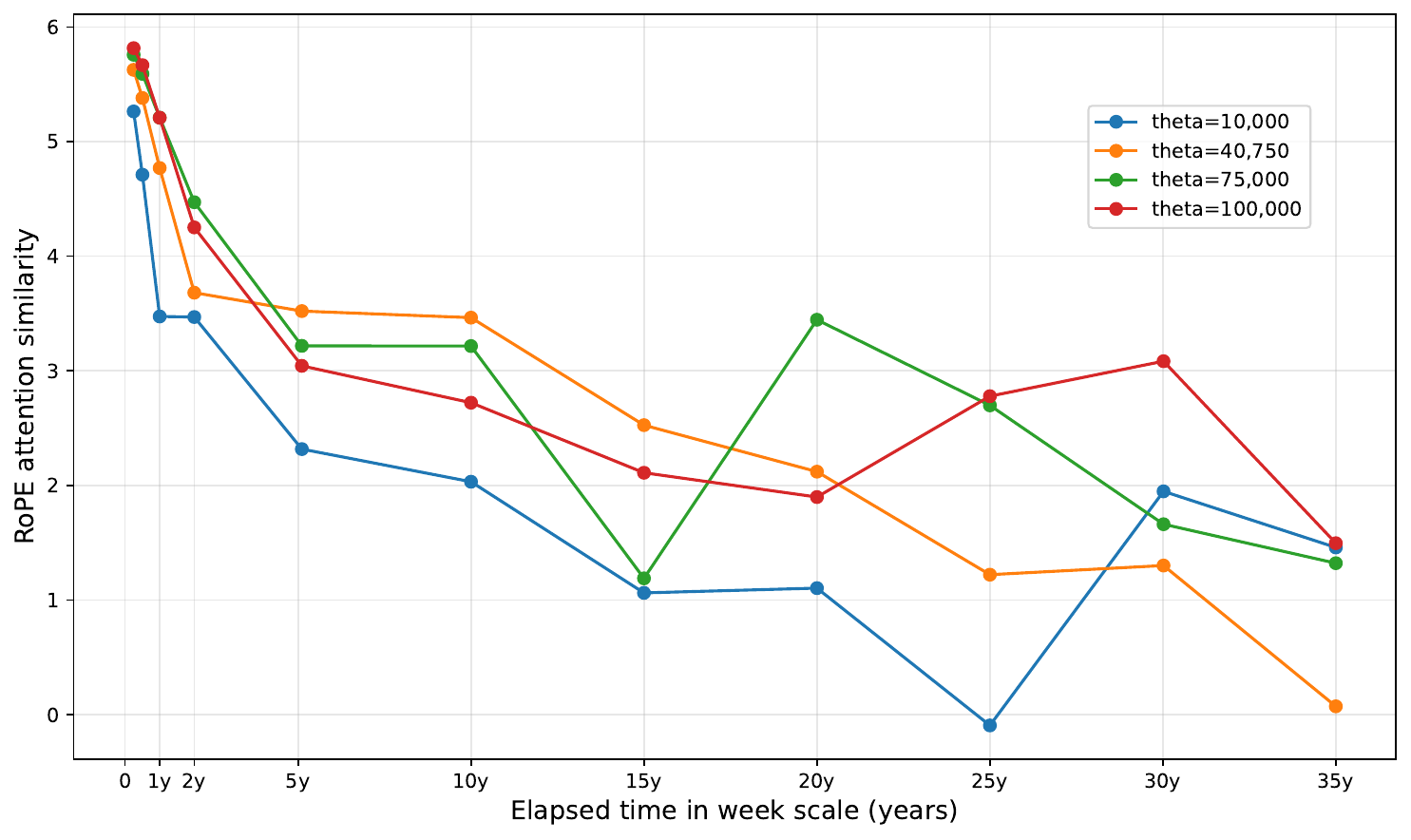}}
\end{figure}

\section{Study design and model figures}\label{apd:figures}
This appendix shows the cohort flowchart (\figureref{fig:flowchart}), the evaluation design (\figureref{fig:design}), and the event representation and model (\figureref{fig:eventstream}).

\begin{figure*}[htbp]
\floatconts
  {fig:flowchart}
  {\caption{Derivation of the pretraining, fine-tuning, validation and evaluation cohorts from the WSIC extract. Controls were matched to cases within each partition.}}
  {\includegraphics[width=0.80\textwidth]{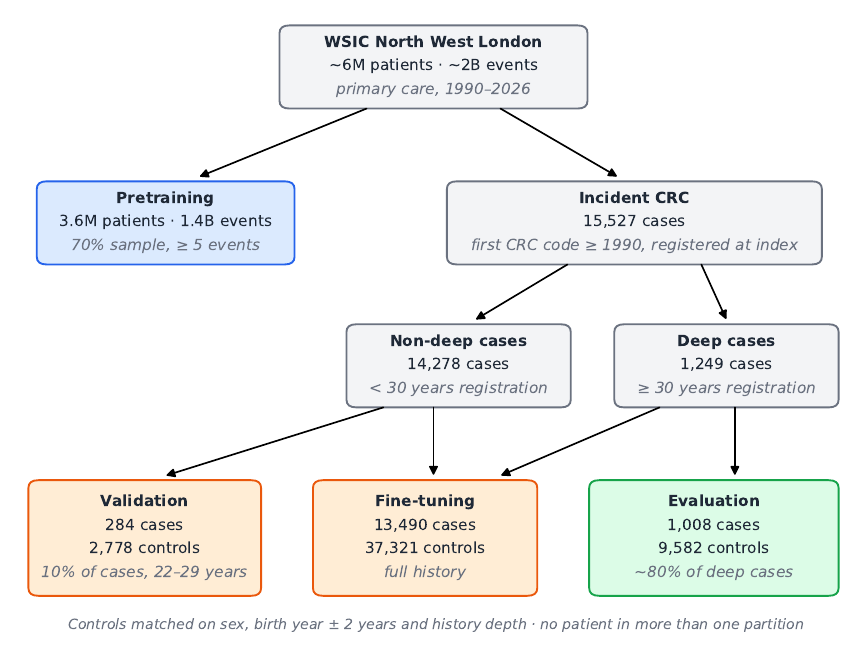}}
\end{figure*}

\begin{figure*}[htbp]
\floatconts
  {fig:design}
  {\caption{Nested design with 15 lookback windows applied to the same 10,590 evaluation patients.}}
  {\includegraphics[width=0.80\textwidth]{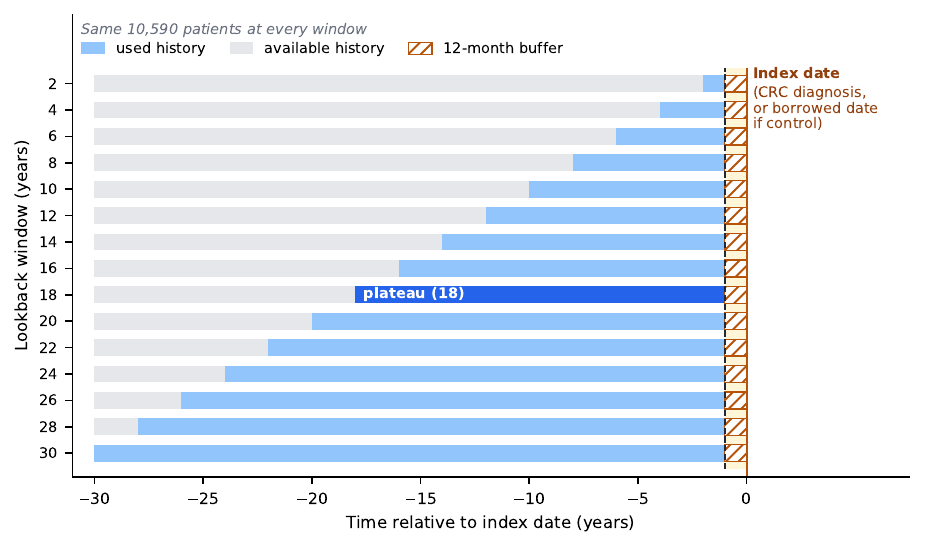}}
\end{figure*}

\begin{figure*}[htbp]
\floatconts
  {fig:eventstream}
  {\caption{Event representation and model. Each event of an illustrative (not real) history is embedded as the sum of code, value and age components; elapsed time between events drives the rotary position embedding. A [CLS] token closes the sequence, and its final hidden state is the only input to the CRC head.}}
  {\includegraphics[width=0.85\textwidth]{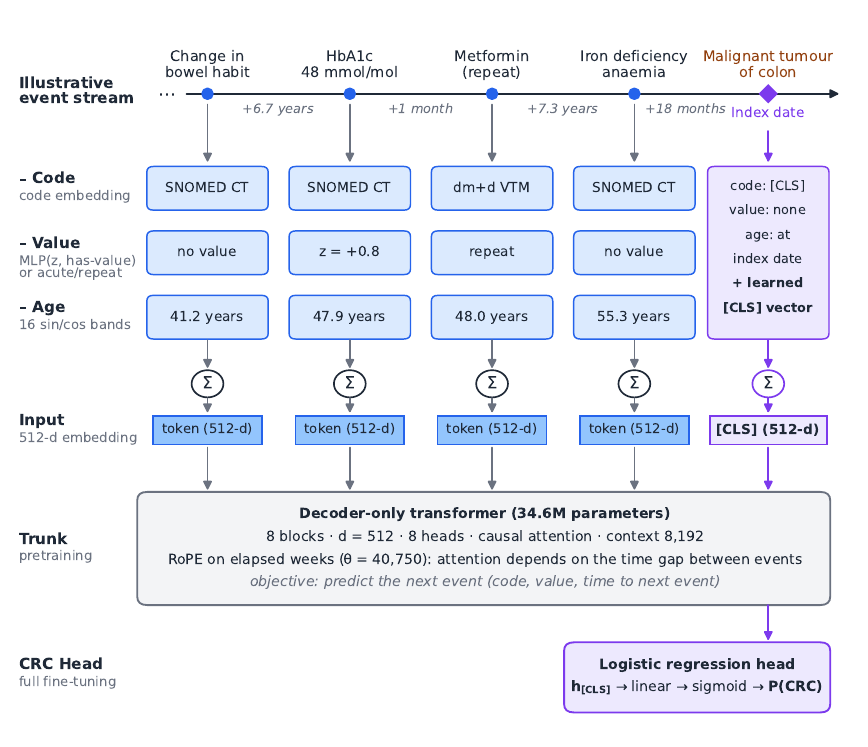}}
\end{figure*}

\section{Data preprocessing}\label{apd:preproc}

Preprocessing ran in SQL on Snowflake within a secure NHS sandbox.

\textbf{Case ascertainment and cohorts.} The first CRC code was identified over the whole record before the 1990 restriction was applied; restricting first would have mistaken a follow-up code for the incident code in 191 prevalent cases.
Under a $\geq$5 events threshold, $\sim$5\% of the cases and $\sim$12\% of the controls were excluded in fine-tuning and validation cohorts, versus 0\% of cases and $\sim$6\% of controls for evaluation. Fine-tuning used 1:3 matched controls as a compromise between class balance and sample size, the loss was reweighted accordingly. Predicted probabilities are therefore not calibrated, and all reported metrics are computed on the separately matched frozen evaluation cohort.

\textbf{Vocabulary curation:} Codes below a frequency floor are mapped to the nearest frequent SNOMED ancestor or dropped (800 patients for findings/procedures; 300 for observable/situations; 50 for dm+d VTMs), and $\sim$2,000 administrative codes are removed on ontological grounds. The four retained hierarchies span 90\% of coded non-prescription events, with 99.98\% within-hierarchy coverage; 91.2\% of prescriptions resolve to a dm+d VTM code. We harmonised dm+d drug concepts across the pre/post-2018 re-identification onto SNOMED product identifiers, removing coding artefacts (e.g., molecule codes splitting or fluctuating across the 2018 boundary) that would otherwise confound deep windows. After unit harmonisation, 96\% of numeric values are retained; missing values use a learned embedding, and prescriptions carry an acute/repeat qualifier.

Within-day repetitions of the same code (order not recoverable) are collapsed to one event, with median retained if value-bearing, reducing the stream by 25\% and avoiding a known source of inflated EHR metrics \citep{rajamohan2025foundation,wornow2025context}.

One evaluation patient exceeded the 8,192-token context and was left-truncated.

\begin{table}[htbp]
\floatconts
  {tab:cohorts}
  {\caption{Cohorts used for pretraining, fine-tuning, model selection and evaluation, after the five-event threshold (number of windows).}}
  {\small\setlength{\tabcolsep}{3pt}
  \begin{tabular}{lrrrl}
  \toprule
  Cohort & Cases & Controls & Ratio & Input\\
  \midrule
  Pretraining & \multicolumn{2}{l}{\textit{3.6M patients}} & --& Full\\
  Fine-tuning & 13,490 & 37,321 & 1:2.77 & Full\\
  Validation & 284 & 2,778 & 1:9.78 & 2--22y (6)\\
  Evaluation & 1,008 & 9,582 & 1:9.51 & 2--30y (15)\\
  \bottomrule
  \end{tabular}}
\end{table}

\section{Model training specifications}\label{apd:training}

All training ran on a single NVIDIA T4 GPU (16 GB) in PyTorch, with fp16 and seed 43 throughout.

\textbf{Pretraining.} Next-event cross-entropy over the 18,733-token vocabulary (output weights tied to the input embedding), with Gaussian negative log-likelihood for values where present and regression on log time-to-next-event, the two auxiliary terms weighted 0.1 each. (1) 40,000 steps on the first half of the pretraining sample: sequences up to 8,192 tokens, batches formed to a fixed budget of 16,384 tokens with length bucketing, AdamW (lr $4.5\times10^{-4}$, weight decay 0.1, $\beta$ = (0.9, 0.95)), cosine schedule with 2,000-step linear warmup, gradient clipping 1.0, fp16. (2) Continuation for 6,500 optimiser steps on the remaining non-overlapping half with 8-fold gradient accumulation ($\sim$131,000 tokens per step), re-warmed to lr $4.5\times10^{-4}$ (400-step warmup, cosine decay), fp16. Final token-level cross-entropy 3.78 against $\ln(18{,}733)$ = 9.84 at initialisation.

Fewer than 0.1\% of pretraining patients had a CRC code; these entered no downstream cohort (e.g. prevalent before 1990 or not registered at diagnosis).

\textbf{Fine-tuning.} lr $2\times10^{-5}$ (trunk) and $10^{-3}$ (head), weight decay 0.05, dropout 0.2, gradient clipping 1.0, one warm-up epoch with frozen trunk, loss weighted by the control:case ratio; best mean validation AUROC at epoch 7.

\textbf{XGBoost baseline.} Depth 6, learning rate 0.05, subsample 0.8, colsample-by-tree 0.5, minimum child weight 0.5, scale pos weight set to the realised control:case ratio; up to 1,000 trees with early stopping (50 rounds) on validation AUROC at the deepest validation window (22 years).

\section{Discrimination by calendar-time lookback}\label{apd:table}
Full AUROC and AUPRC results are given in \tableref{tab:auroc-full,tab:auprc-full}.

\begin{table}[!ht]
\floatconts
  {tab:auroc-full}
  {\caption{AUROC (95\% CI) by calendar-time lookback depth for the transformer and the XGBoost baseline on the frozen evaluation cohort ($n = 10{,}590$; 1,008 cases). Confidence intervals are 2.5th--97.5th percentiles of 1,000 patient-level bootstrap replicates. Peaks in bold; for XGBoost, 14 and 24 years differ by 0.0001 and are both shown.}}
  {\footnotesize\setlength{\tabcolsep}{3pt}
  \begin{tabular}{rll}
  \toprule
  Window (y) & Transformer & XGBoost\\
  \midrule
  2  & 0.5505 (0.5320--0.5691) & 0.5281 (0.5101--0.5481)\\
  4  & 0.5625 (0.5433--0.5812) & 0.5445 (0.5266--0.5650)\\
  6  & 0.5730 (0.5530--0.5922) & 0.5553 (0.5359--0.5731)\\
  8  & 0.5756 (0.5555--0.5950) & 0.5625 (0.5437--0.5809)\\
  10 & 0.5825 (0.5637--0.6003) & 0.5718 (0.5533--0.5902)\\
  12 & 0.5840 (0.5643--0.6020) & 0.5795 (0.5596--0.5981)\\
  14 & 0.5814 (0.5615--0.6012) & \textbf{0.5886} (0.5690--0.6067)\\
  16 & 0.5868 (0.5679--0.6048) & 0.5881 (0.5695--0.6066)\\
  18 & \textbf{0.5901} (0.5717--0.6080) & 0.5857 (0.5673--0.6032)\\
  20 & 0.5785 (0.5604--0.5964) & 0.5880 (0.5698--0.6050)\\
  22 & 0.5841 (0.5668--0.6024) & 0.5880 (0.5706--0.6054)\\
  24 & 0.5791 (0.5606--0.5974) & \textbf{0.5887} (0.5706--0.6059)\\
  26 & 0.5787 (0.5600--0.5965) & 0.5863 (0.5689--0.6036)\\
  28 & 0.5807 (0.5622--0.5990) & 0.5848 (0.5671--0.6019)\\
  30 & 0.5802 (0.5612--0.5990) & 0.5844 (0.5661--0.6017)\\
  \bottomrule
  \end{tabular}}
\end{table}

\begin{table}[!ht]
\floatconts
  {tab:auprc-full}
  {\caption{AUPRC (95\% CI) by calendar-time lookback depth, same cohort and bootstrap as \tableref{tab:auroc-full}. AUPRC is reported against the realised evaluation prevalence (0.095) and is not comparable to population incidence. Peaks in bold.}}
  {\footnotesize\setlength{\tabcolsep}{3pt}
  \begin{tabular}{rll}
  \toprule
  Window (y) & Transformer & XGBoost\\
  \midrule
  2  & 0.1128 (0.1026--0.1245) & 0.1082 (0.0982--0.1200)\\
  4  & 0.1169 (0.1062--0.1289) & 0.1133 (0.1022--0.1261)\\
  6  & 0.1206 (0.1098--0.1323) & 0.1162 (0.1060--0.1289)\\
  8  & 0.1239 (0.1118--0.1367) & 0.1176 (0.1074--0.1308)\\
  10 & 0.1259 (0.1141--0.1385) & 0.1193 (0.1093--0.1322)\\
  12 & 0.1265 (0.1144--0.1393) & 0.1225 (0.1120--0.1346)\\
  14 & 0.1272 (0.1153--0.1406) & \textbf{0.1256} (0.1151--0.1380)\\
  16 & 0.1293 (0.1173--0.1430) & 0.1252 (0.1148--0.1381)\\
  18 & \textbf{0.1323} (0.1197--0.1469) & 0.1235 (0.1133--0.1361)\\
  20 & 0.1270 (0.1157--0.1410) & 0.1244 (0.1136--0.1367)\\
  22 & 0.1278 (0.1163--0.1420) & 0.1242 (0.1141--0.1368)\\
  24 & 0.1284 (0.1164--0.1425) & 0.1243 (0.1143--0.1362)\\
  26 & 0.1273 (0.1153--0.1414) & 0.1240 (0.1142--0.1362)\\
  28 & 0.1285 (0.1169--0.1431) & 0.1245 (0.1144--0.1367)\\
  30 & 0.1284 (0.1166--0.1441) & 0.1239 (0.1141--0.1363)\\
  \bottomrule
  \end{tabular}}
\end{table}

\end{document}